\def\year{2019}\relax
\documentclass[letterpaper]{article} 
\usepackage{aaai19}  
\usepackage{times}  
\usepackage{helvet} 
\usepackage{courier}  
\usepackage[hyphens]{url}  
\usepackage{graphicx} 
\usepackage{graphicx}  
\title{Macro Action Selection with Deep Reinforcement Learning in StarCraft}

\author{
Sijia Xu,\textsuperscript{\rm 1}
Hongyu Kuang,\textsuperscript{\rm 2}
Zhuang Zhi,\textsuperscript{\rm 1}
Renjie Hu,\textsuperscript{\rm 1}
Yang Liu,\textsuperscript{\rm 1}
Huyang Sun\textsuperscript{\rm 1}\\
\textsuperscript{\rm 1}Bilibili\\
\textsuperscript{\rm 2}State Key Lab for Novel Software Technology, Nanjing University\\
\{xusijia, zhuangzhi, hurenjie, liuyang01, sunhuyang\}@bilibili.com, khy@nju.edu.cn
}

\begin{document}

\maketitle
\begin{abstract}
StarCraft (SC) is one of the most popular and successful Real Time Strategy (RTS) games. In recent years, SC is also widely accepted as a challenging testbed for AI research because of its enormous state space, partially observed information, multi-agent collaboration, and so on. With the help of annual AIIDE and CIG competitions, a growing number of SC bots are proposed and continuously improved. However, a large gap remains between the top-level bot and the professional human player. One vital reason is that current SC bots mainly rely on predefined rules to select macro actions during their games. These rules are not scalable and efficient enough to cope with the enormous yet partially observed state space in the game. In this paper, we propose a deep reinforcement learning (DRL) framework to improve the selection of macro actions. Our framework is based on the combination of the Ape-X DQN and the Long-Short-Term-Memory (LSTM). We use this framework to build our bot, named as LastOrder. Our evaluation, based on training against all bots from the AIIDE 2017 StarCraft AI competition set, shows that LastOrder achieves an 83\% winning rate, outperforming 26 bots in total 28 entrants.
\end{abstract}

\section{Introduction}

\begin{figure}
	\center
	\includegraphics[width=.95\columnwidth]{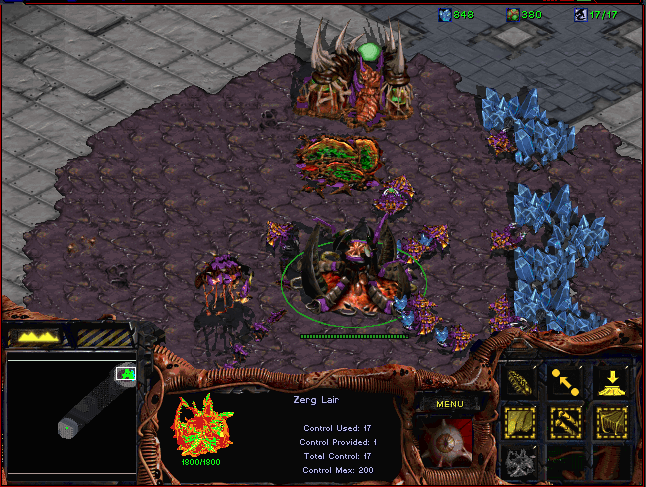}
	\caption{A screenshot of StarCraft: Brood War}
	\label{fig:screenshot}
\end{figure}

StarCraft: Brood War (SC) is one of the most popular and successful Real Time Strategy (RTS) games created by Blizzard Entertainment in 1998.
Under the setting of a science-fiction based universe, the player of SC picks up one of the three races: Terran, Protoss, or Zerg, to defeat other players in a chosen map. Figure~\ref{fig:screenshot} presents a screenshot of SC showing a play using the Zerg race.
In general, to achieve victory in a standard SC game, the player needs to perform a variety of actions, including gathering resources, producing units, updating technologies and attacking enemy units.
These actions can be categorized into two basic types: micro actions (Micro) and macro actions (Macro)~\cite{ontanon2013survey}:

\textbf{Micro}. The micro actions manipulate units to perform operation-level tasks such as exploring regions in a map, collecting resources and attacking the enemy units.
The general goals of micro actions during the entire game are: (1) to keep units performing more tasks; (2) to avoid being eliminated by the enemy.

\textbf{Macro}. The macro actions represent the strategy-level planning to compete with the opponent in the game, such as the production of combat units, the placement of different buildings and the decision of an attack.
The general goal of macro actions is to efficiently counter the opponent's macro actions throughout the game.
It is worth noting that, in a standard SC game, regions in the map that are not occupied by the player's units or buildings are kept unknown. The so-called ``fog of war'' mechanism leads to partial observations on the map and thus significantly increases the difficulty to select proper macro actions.

In recent years, SC has been widely accepted as a challenging testbed for AI researchers, because of the multi-unit collaboration in the Micro and the decision-making on enormous state spaces in the Macro.
A growing number of bots based on different AI techniques are proposed and continuously improved, especially in annual StarCraft AI competitions held by both AAAI Conference on Artificial Intelligence and Interactive Digital Entertainment (AIIDE), and IEEE Conference on Computational Intelligence and Games (CIG).
Unfortunately, although the ability of SC bots has been greatly improved, a large gap remains between the top-level bot and the professional human player.
For example, in the game ladder iCCup where players and bots are ranked by their SC match results~\cite{ontanon2013survey}, the best SC bot ranks from D to D+. On the contrary, the average amateur player ranks from C+ to B, while the professional player usually ranks from A- to A+. One vital reason why bots fall behind human players is that current bots mainly rely on predefined rules to perform macro actions. These rules are not scalable and efficient enough to cope with the enormous yet partially observed state space in the game.

Recently, DRL-based bots have achieved substantial progress in a wide range of games, such as Atari games~\cite{mnih2015human}, Go~\cite{silver2017mastering}, Doom~\cite{wu2016training} and Dota2~\cite{openaifive}.
However, two main challenges remain when using the DRL framework to perform better macro actions in SC.
The first one is the partial observation problem caused by the fog of war in the map. This problem makes the current observations insufficient to infer the future states and rewards.
The other challenge is the sparse reward problem, for example, in SC the length of a usual game is around 20 minutes with 250-400 macro actions performed,
thus it is hard to get positive rewards using only the terminal reward at the end of the game when training from scratch.

To address these two challenges, we combine the reinforcement learning approach Ape-X DQN~\cite{horgan2018distributed} with the Long-Short-Term-Memory model (LSTM)~\cite{hochreiter1997long} to propose our DRL-based framework. Specifically, the LSTM is used to address the partial observation problem, and the Ape-X is used to address the sparse reward problem. Then, we use this framework to build our bot, named as LastOrder\footnote{LastOrder is the name of an immature AI character in a popular animated drama called Toaru Majutsu no Index}. By training against all bots from the AIIDE 2017 StarCraft AI Competition, LastOrder achieves an overall 83\% win-rate, outperforming 26 bots in total 28 entrants. Furthermore, the same version of LastOrder attends the AIIDE 2018 StarCraft Competition and ranks 11 in total 25 entrants. We have also open-sourced LastOrder with the training framework at https://github.com/Bilibili/LastOrder.

\section{Related Work}

Researchers have focused on macro action selection in StarCraft for a long time~\cite{ontanon2013survey}. One way used by many bots, e.g., UAlbertaBot~\cite{UALBOT} and AIUR~\cite{AIUR}, is the bandit learning method based on several predefined macro action sequences. Bandit learning can choose the most appropriate macro action sequence using the historical match. However, In SC, typical bandit learning method is only used at the start of the game. Thus, it's insufficient to cope with the large state space throughout the game.

Meanwhile, a different way to solve this problem is the data mining from human players' replays. Hsieh et al. propose to learn actions from a large number of replays~\cite{hsieh2008building}. They learn every detailed click from human players. Weber et al. encode the whole game into a vector for each player and then model the problem into a supervised learning problem~\cite{weber2009data}. Kim et al. propose to categorize different building orders and summary them into separate states and actions~\cite{kim2010cooperative}. Different from the above work, Hostetler et al. start to consider partially observe problem in RTS game by using dynamic Bayes model inference~\cite{hostetler2012inferring}. Many researchers~\cite{synnaeve2012dataset}~\cite{robertson2014improved} also propose datasets of replays which contain much richer information. However, for full bot development, replay mining method remains unsatisfying for two reasons. First, the macro action sequence from professional players may not be the best choice for the bots, due to the large difference in micro management ability between bots and professional players. Second, some tactic-level macro actions, e.g., when to trigger a specific kind of attack to a destination, are important to the overall performance of bots. But it is unlikely for replay mining method to extract these highly abstracted tactic-level actions.

In recent years, DRL-based methods have achieved noticeable success in building autonomous agents. Currently, there are mainly two ways to apply DRL to RTS games. The first one is applying DRL to the micro management. Kong et al.~\cite{kong2017revisiting}~\cite{kongeffective} propose several master-slave and multi-agent models to help controlling each unit separately in specific scenario. Shao et al.~\cite{shao2018starcraft} introduce transfer learning and sharing multi-agent gradient to train cooperative behaviors between units. These proposed methods are capable of performing excellent work in a small combat scenario, but it remains hard to scale to a large combat scenario and to react instantly at the same time. This limitation also restricts the practical application of DRL based micro management in full bot development. The other way is applying DRL to the macro management. Compared with replay-mining methods, macro actions learned through DRL can directly match the bot's micro management ability. Furthermore, macro actions learned from DRL can include both native macro actions (e.g., building, producing, and upgrading) and customized tactic-level macro actions. Sun et al.~\cite{sun2018tstarbots} created a StarCraft II bot based on a DRL framework to do macro actions selection, and achieved desirable results when competing with the build-in Zerg AI. By contrast, we focus more on handling partial observation and sparse reward problems to compete against a wider range of bots.

\section{Proposed Method}
\subsection{Actions, state and reward}
\subsubsection{Macro actions}~\\ 
We define 54 macro actions for Zerg covering the production, building, upgrading and different kinds of attack as summarized in Table~\ref{tab:macro_action_category}(for the full list please refer to our project on GitHub).
All Macro actions excluding the attack actions have direct meaning in StarCraft. For attack actions, each one represents a attack with a specific mode (e.g., harassing enemy natural base using mutalisks).

\begin{table}
	\centering
	\caption{Summary of 54 macro actions}
	\label{tab:macro_action_category}
		\begin{tabular}{lll}
        \hline
		Category & Actions & Examples \\
        \hline
		Production & 7 & ProduceZergling \\
		Building & 14 & BuildLair \\
		Upgrading & 15 & UpgradeZerglingsSpeed \\
		Attack & 17 & AttackMutaliskHarassNatural \\
		Expansion & 1 & ExpandBase  \\
		Waiting & 1 & WaitDoNothing \\
		\hline
	\end{tabular}
\end{table}

\subsubsection{Micro actions}~\\ 
We define a large set of detailed micro actions to manipulate different units and buildings of the Zerg race. In general, these actions manipulate units and buildings to perform operation-level tasks, such as moving to somewhere and attacking other units. 

\subsubsection{State}~\\ 
The state come as a set of current observation features and history features.
Current observation features describe the current status of ours and enemies, e.g., units and buildings.
History features are mainly designed to keep and accumulate enemy information from the start of the game. For example, once we observe a new enemy building, unless it is destroyed by us, we add a feature describing its existence whether the enemy building is under the fog of war.

\subsubsection{Reward}~\\ 
Reward shaping is an effective technique to reinforcement learning in a complicated environment with delayed reward~\cite{ng1999policy}. In our case, instead of using a terminal result (1(win)/-1(loss)), we use a modified terminal reward with in-game score as below (where \emph{timedecay} refers to the game time):

\begin{equation}
r_{new}=\gamma^{timedecay}\frac{score_{our} - score_{enemy}}{\max{(score_{our}, score_{enemy})}}
\end{equation}

The in-game score is defined by the SC game engine including the a building score, a unit score and a resource score to reflect the player’s overall performance. The modified terminal score describes the quality of the terminal result. We find that it can guide the exploration more efficiently. For example, policy in a game with a higher modified terminal score is better than others even though games are all lose.

\subsection{Learning Algorithms and Network Architectures}
\subsubsection{Ape-X DQN}~\\
In SC, it is hard to get positive reward using only the terminal reward at the end of a game when training from scratch.
This sparse reward problem become severe when training against strong opponent.
Recently, a scaled up variant DQN called Ape-X DQN~\cite{horgan2018distributed} achieves a new state of arts performance on Atari games, especially on some well-known sparse reward game like Montezuma’s Revenge.
They suggest that using a large number of actors may help to discover new courses of policy and avoid the local optimum problem caused by insufficient exploration.
This scaled up approach is a relatively easy way to solve the sparse reward problem.

Specifically, Ape-X DQN uses double Q-learning, multi-step bootstrap targets, prioritized replay buffer and duel network. In our case, there is no instant reward, the loss function is $l_{t}(\theta)=\frac{1}{2}(G_{t}-q(S_{t}, A_{t}, \theta))^{2} $ with the following definition (where \emph{t} is a time index for the sample from the replay starting with state $S_{t}$ and action $A_{t}$, and $\theta^{-}$ denotes parameters of the target network):

\begin{equation}
G_{t}=\lambda^{n}q(S_{t+n}, \mathop{argmax}_{a} q(S_{t+n}, a, \theta), \theta^{-})
\end{equation}

Although it is uncommon to use multi-step bootstrap targets in off-policy learning without off-policy correction, we find that this setting performs better than single step target setting. We suggest that using low exploration rate on majority of actors may improve the on-policy degree of training data, and the performance of multi-step bootstrap targets is also improved in off-policy setting with the on-policy training data. Besides, in our case, instead of using a center replay memory to store all transitions in FIFO order, we split the replay memory into multiple segments which equal to the number of opponents, due to the unbalanced training transition generating speed of different opponents. Each opponent only update transition on its own replay memory segment in FIFO order. During the training stage, transition is sampled over all replay memory segments according to its prioritization.

\subsubsection{Deep Recurrent Q-Networks}~\\
In SC, the macro action selection decision process is non-Markovian, because the existence of fog of war causes the future states and rewards depending on more than current observation. The macro action selection process becomes a Partially Observable Markov decision process (POMDP) and the DQN’s performance declines when given incomplete state observations. To deal with this problem, Hausknecht and Stone~\cite{hausknecht2015deep} introduce the Deep Recurrent Q-Networks (DRQN). According to their experiment, adding a LSTM layer to the normal DQN model can approximate more accurate Q-values, leading to better policies in partially observed environments.

In our case, we use the Ape-X DQN instead of the normal DQN to achieve better performance. Besides, in order to cover longer time horizon, the interval between each LSTM step is extended to 10 seconds. We observed that a 10-seconds interval is short enough to reflect changes in macro state without missing too much detail changes.

\subsection{Training}
We use 1000 machines(actor) to run 500 parallel games against different opponents scheduled by StarCraftAITounrnamentManger~\cite{TM}.
Similar to TorchCraft~\cite{synnaeve2016torchcraft}, there are two parts in each actor. The model part uses a separate python process to handle macro action selection based on TensorFlow~\cite{abadi2016tensorflow}, and the other part is a BWAPI~\cite{bwapi} DLL performing all actions in SC. These two parts use a message queue to exchange message between each other.

During each game, we cache observations in the actor's memory and group them into observation sequences.
To alleviate the load of leaner which receives transitions, we only send transitions at the end of the game.
The generation speed of transition is approximately 20000 per minute. The learner's update speed is about 10 batches (196 batch size) per second. Actors copy the network parameters from the learner every 10 seconds. Each actor $i\in\{0,1,…,N-1\}$ executes an $\epsilon_{i}$ greedy policy where $\epsilon_{i}=\epsilon^{1+\frac{i}{N-1}\alpha} $ with $\epsilon=0.4,\alpha=7,N=1000$.
The capacity of each replay memory segment is limited to one million transitions. Transition is sampled over all replay memory segments according to its proportional prioritization with a priority exponent of 0.6 and an importance sampling exponent set to 0.4.

\section{Experiment}
\subsection{A controlled case for analysis}

For qualitative analysis we train our bot against six selected bots as described in Table~\ref{tab:description_opponent} on a single map.
We select Ximp bot and cpac bot to show the sparse reward problem. Both of them use the mid-game rush strategy with stable early game defense and are relatively difficult to defeat.
The rest of bots are added to intensify the partial observation problem, because the opponents in the same race usually have some similar macro states during the game.

\begin{figure}
	\center
	\includegraphics[width=.95\columnwidth]{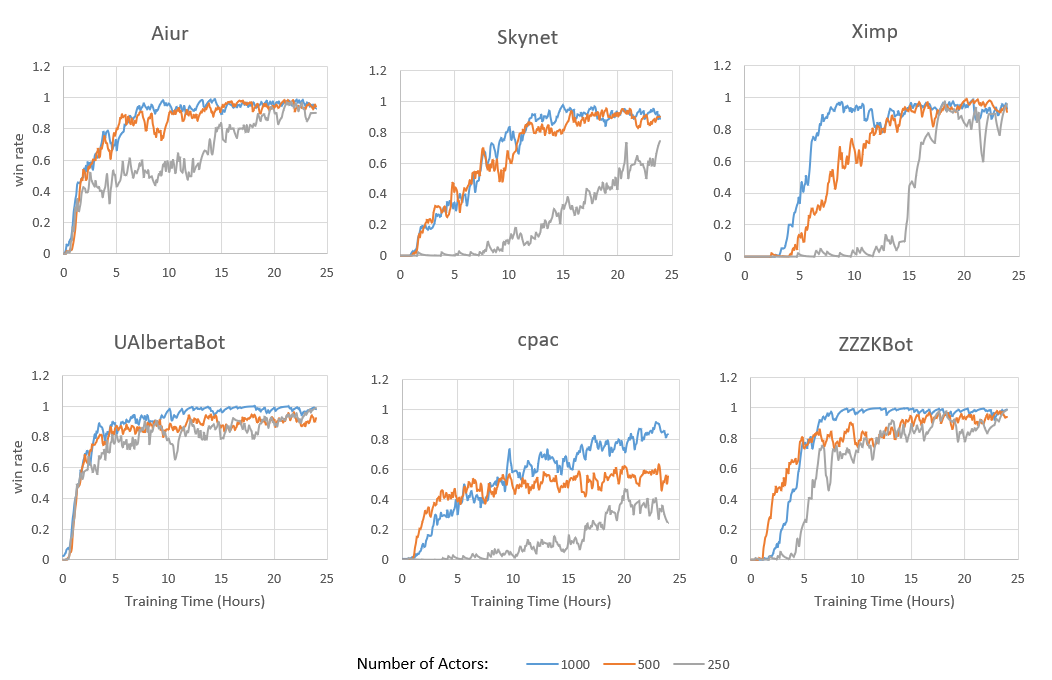}
	\caption{Performance consistently improves as the number of actors increased from 250 to 1000.}
	\label{fig:actor_num}
\end{figure}

In Figure~\ref{fig:actor_num}, we show the learning curve of win-rate against each bot with a different number of actors. The win-rate are evaluated by actors with zero exploration rate. It is worth noting that with 1000 actors almost all the win-rate become close to 1 after 24 hours' training.

\begin{table}
	\centering
	\caption{Six selected opponents (the ranking is based on AIIDE 2017)}
	\label{tab:description_opponent}
	\begin{tabular}{llll}
		\hline
		Bot name & Rank & Race & Strategy \\
		\hline
		ZZZKBot & 1 & Zerg & Early rush \\
		
		cpac & 4 & Zerg & Mid-game \\
		
		Ximp & 13 & Protoss & All-in Carrier rush \\
		
		UAlbertaBot & 14 & Random & 9 strategies on 3 races \\
		
		Aiur & 15 & Protoss & 4 different strategies \\
		
		Skynet & 17 & Protoss & 3 different strategies \\
		\hline
	\end{tabular}
\end{table}

\subsubsection{A detailed case: LastOrder vs. Ximp}~\\

We then use Ximp bot for case analysis because the sparse reward problem in Ximp is severest among the six bots. Ximp's early game strategy is to do stable defense with many defense buildings and little combat units. The increasing number of defense buildings and limited combat units are gradually different from other Protoss bots.
The mid game strategy of Ximp is to produce top-level combat unit (Carrier) as soon as possible to attack opponent in order to gain big advantage or directly win the game in one shot. Compared with other opponents' counterstrategy, the length of counterstrategy against Ximp is much longer, because LastOrder must first defend Carrier rush in the mid game and then defeat Ximp in the later game. Thus LastOrder needs much more exploration effort in order to get the positive reward when training from scratch.

In Figure~\ref{fig:actor_num}, after 24 hours' training with 1000 actors, due to the lack of harassing strategy in Ximp, in the early game, LastOrder's counterstrategy is to quickly expand multiple bases and produce workers in order to gain big economy advantage. This is totally different from the learned counterstrategies against other Protoss bots, for example, the strategy against Skynet in the early game is to build defense buildings and to produce combat units in order to counter its early rush.

In the mid game, LastOrder finds the efficient countering battle unit (Scourage) and produce a sufficient number of Scourages to defeat Ximp's Carrier attack, the choice of producing Scourages is also different from counterstrategies of other bots. Due to the big economy advantage in the early game, in the later game LastOrder has sufficient resource to produce other combat units along with the Scourages. Thus it is relatively easy to win the game in the end.

\subsubsection{Component analysis}~\\

We also run additional experiments to improve our understanding of the framework and to investigate the contribution of different components.
First, we investigate how the performance scale with the number of actors. We trained our bot with different number of actors (250, 500, 1000) for 24 hours against the 6 opponents set as described in Table~\ref{tab:description_opponent}. Figure~\ref{fig:actor_num} shows that the performance consistently improved as the number of actors increased without changing any hyper-parameter or the structure of the network. This is also similar to the conclusion in Ape-X DQN.

\begin{figure}
	\center
	\includegraphics[width=.95\columnwidth]{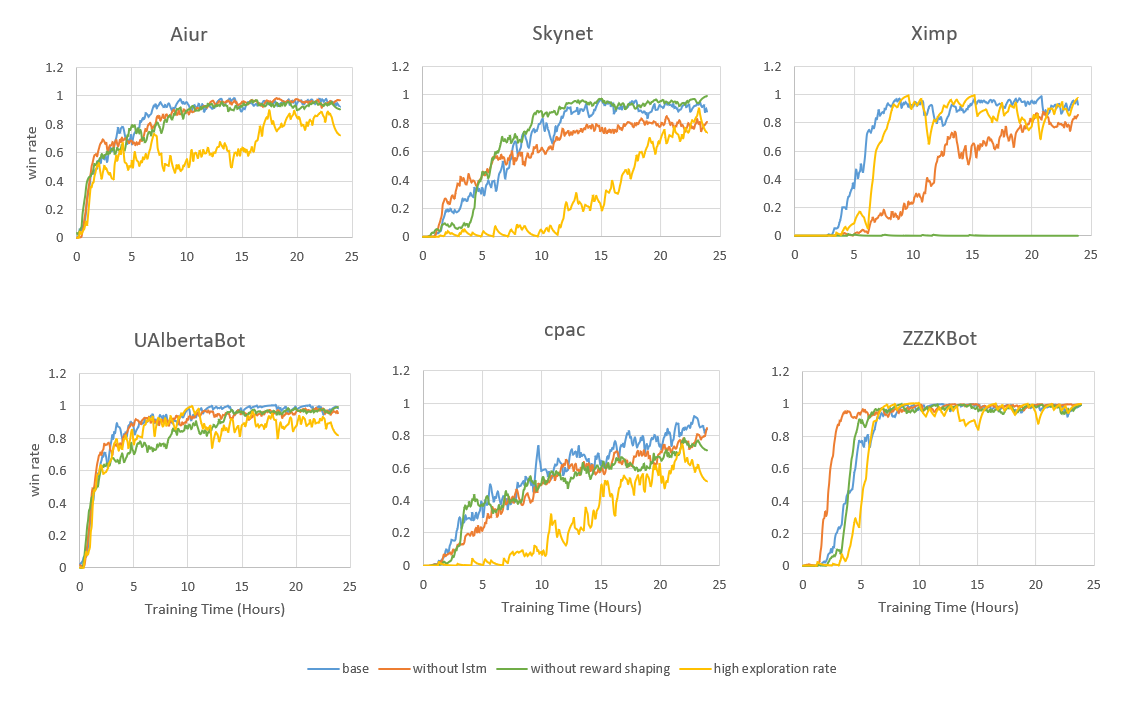}
	\caption{Four different experiments. Each experiment using 1000 actors trained for 24 hours.}
	\label{fig:comapre}
\end{figure}

Next, In Figure~\ref{fig:comapre}, we run three experiments to investigate the influence of other components:
\begin{itemize}
	\item \textbf{Without LSTM.} We only use the latest observation instead of a period of observations to investigate the influence of partial observation. The experiment shows a small performance drop against Protoss bots (Skynet and Ximp). We suggest that model is less likely to differentiate states among the same race without the sequence of observation.
	
	\item \textbf{Without reward shaping.} We use the origin 1(win)/-1(loss) as the terminal reward instead of the modified terminal reward. According to this experiment, the win-rate of Ximp is kept at about 0\% over the whole training time. We suggest that the exploration is much harder to get improved reward without the help of reward shaping to alleviate the sparse reward problem, especially when training against strong opponent.
	
	\item \textbf{High exploration rate.} In this experiment, each actor executes an $\epsilon_{i}$ greedy policy where $\epsilon_{i}=\epsilon^{1+\frac{i}{N-1}\alpha} $ with $\epsilon=0.7,\alpha=11,N=1000$. This setting corresponds to a higher exploration rate in actors. The overall performance declines in many bots (cpac, Aiur, Skynet) compared with the low exploration rate setting. We suggest that using low exploration rate on majority of actors is equal to getting on-policy transitions with relatively little noisy. As the learning proceeds, the change in policy become less and the transitions in replay buffer become more on-policy. This combination of low exploration on majority of actors and multi-step bootstrap targets improve the overall performance.
\end{itemize}

\subsection{Evaluation on AIIDE 2017 StarCraft AI Competition bots set}
\label{sec:AIIDE2017}
We then train our bot against the AIIDE 2017 StarCraft AI Competition bots set over 10 maps to show the performance and scalability.
The 2017 bots set is comprised of 28 opponents including 4 Terran bots, 10 Protoss bots and 14 Zerg bots. With the need of parallel training and fast evaluation, we do not use the round robin mode in StarCraftAITounrnamentManger~\cite{TM}. All games are created as soon as possible without waiting for previous round to finish. This running mode also disable the opponent's online learning strategy and may cause potential performance drop in LastOrder in round robin mode.
The final trained LastOrder achieves 83\% win-rate in 8000 games, outperforming 26 bots on 28 bots set. The detailed evaluation result is listed in Table ~\ref{tab:evaluation}.

\subsection{AIIDE 2018 StarCraft AI Competition}
We use the pre-trained LastOrder in~\ref{sec:AIIDE2017} to attend the AIIDE 2018 StarCraft Competition and rank 11 in total 25 entrants. The official result can be found in~\cite{AIIDE2018}.

The performance drop has two reasons. First, when the opponent's micro management suppress ours, our enhancement to macro actions is insufficient to win the game. This happens for SAIDA, CherryPi, Locutus, two Locutus-based bots (CSE and DaQin), and McRave in 2018 competition. Second, The insufficient Terran opponents in the training sets(only 4 Terran bots in the 2017 bots set) leads to low win-rate when playing against new Terran bots Ecgberht and WillyT.

Despite the discussed two reasons, on the rest of newly submitted bots set, LastOrder achieves about 75\% win-rate. Considering the use of online learning strategy by other bots in round robin mode, this win-rate is close to the offline evaluation result.

\section{Conclusions and Future Work}
Developing a strong bot to act properly in StarCraft is a very challenging task. In this paper, we propose a new framework based on DRL to improve the performance  of the bot's macro action selection. Via playing against AIIDE 2017 StarCraft Competition bots, our bot achieve 83\% win-rate showing promising result.

Our future work involves the following three aspects:

\textbf{Fix bug and optimize micro.} Current macro action execution and micro management is hard coded. The bug in these codes may cause unusual variance in transitions, and it may severely influence training. Besides, the quality of micro is also a key aspect to the performance of bot. e.g., Iron bot used to place building to block the choke point, whereas current LastOrder can not identify the blocked choke point. This usually results in a huge army loss and the macro model cannot help with it.

\textbf{Self-play training.} We observe that to some extent the performance of model is highly constrained by training opponents. If opponents have bugs like a group of armies stuck somewhere, even though we can get a high reward in the end, it is not a valuable reward and may lead the policy to a wrong direction. A better solution may be the self-play training. But the self-play training needs micro and macro codes of three races which may also be a big overhead.

\textbf{Unit level control.} Micro management in StarCraft is a multi-agent system. It is difficult for rule-controlled units to behave properly in different situations and cooperate with each other. But how to train this multi-agent model and react in real time at a relatively large scale is still an open question.

\begin{table}[t]
	\centering
	\caption{The detailed evaluation result of LastOrder against AIIDE 2017 StarCraft AI competition bots without IO sync.}
	\label{tab:evaluation}
	\begin{tabular}{llll}
		\hline
		Bot name & Race & Games & Win rate \% \\
		\hline
		Iron & Terran & 303 & 98.68  \\
		
		PurpleWave & Protoss & 303 & 97.35  \\
		
		LastOrder & Zerg & 8484 & 83.06  \\
		
		Microwave & Zerg & 303 & 49.67  \\
		
		Ximp & Protoss & 303 & 31.91  \\
		
		LetaBot & Terran & 303 & 28.48  \\
		
		IceBot & Terran & 303 & 22.77  \\
		
		Arrakhammer & Zerg & 303 & 15.18  \\
		
		Skynet & Protoss & 303 & 15.13  \\
		
		Juno & Protoss & 303 & 15.13  \\
		
		Steamhammer & Zerg & 303 & 13.82  \\
		
		cpac & Zerg & 303 & 12.5  \\
		
		AILien & Zerg & 303 & 12.21  \\
		
		UAlbertaBot & Random & 303 & 10.6  \\
		
		McRave & Protoss & 303 & 9.54  \\
		
		Myscbot & Protoss & 303 & 9.24  \\
		
		CherryPi & Zerg & 303 & 7.59  \\
		
		Overkill & Zerg & 303 & 6.27  \\
		
		Aiur & Protoss & 303 & 3.29  \\
		
		Xelnaga & Protoss & 303 & 2.96  \\
		
		KillAll & Zerg & 303 & 1.97  \\
		
		GarmBot & Zerg & 303 & 1.65  \\
		
		Tyr & Protoss & 303 & 1.65  \\
		
		MegaBot & Protoss & 303 & 1.64  \\
		
		ZZZKBot & Zerg & 303 & 1.64  \\
		
		Sling & Zerg & 303 & 1  \\
		
		TerranUAB & Terran & 303 & 0.99  \\
	
		Ziabot & Zerg & 303 & 0.99  \\
		
		ForceBot & Zerg & 303 & 0.66  \\
		\hline
	\end{tabular}
\end{table}

\bibliography{cite}
\bibliographystyle{aaai}

\end{document}